\documentclass[preprint,12pt]{elsarticle}

\usepackage{amssymb}
\usepackage{amsmath}
\usepackage{url} 
\usepackage{textcomp}
\usepackage{graphicx}%
\usepackage{multirow}%
\usepackage{amsmath,amssymb,amsfonts}%
\usepackage{amsthm}%
\usepackage{mathrsfs}%
\usepackage{xcolor}%
\usepackage{textcomp}%
\usepackage{manyfoot}%
\usepackage{booktabs}%
\usepackage{algorithm}%
\usepackage{algorithmicx}%
\usepackage{algpseudocode}%
\usepackage{listings}%
\usepackage{amssymb}
\usepackage[figuresright]{rotating}
\usepackage{CJKutf8}
\usepackage{CJK}
\usepackage{makecell}
\usepackage{verbatim}
\usepackage{subfigure}
\usepackage{bm}
\usepackage{tabularx}
\usepackage{adjustbox}

\journal{Knowledge-Based Systems}

\begin{document}

\begin{frontmatter}



\title{A BERT-based Hierarchical Classification Model with Applications in Chinese Commodity Classification}


\author{Kun Liu, Tuozhen Liu, Feifei Wang, and Rui Pan}

\affiliation{organization={School of Statistics and Mathematics, Central University of Finance and Economics},
            addressline={2023210967@email.cufe.edu.cn}, 
            city={Beijing},
            postcode={100081}, 
            state={Beijing},
            country={China}}

\affiliation{organization={Reigning Capital Co., Ltd.},
            addressline={liutuozhen@reigning-capital.com}, 
            city={Beijing},
            postcode={100080}, 
            state={Beijing},
            country={China}}

\affiliation{organization={Center for Applied Statistics and School of Statistics, Renmin University of China},
            addressline={feifei.wang@ruc.edu.cn}, 
            city={Beijing},
            postcode={100872}, 
            state={Beijing},
            country={China}}
            
\affiliation{organization={School of Statistics and Mathematics, Central University of Finance and Economics},
            addressline={ruipan@cufe.edu.cn}, 
            city={Beijing},
            postcode={100081}, 
            state={Beijing},
            country={China}}

\begin{abstract}
Existing e-commerce platforms heavily rely on manual annotation for product categorization, which is inefficient and inconsistent. These platforms often employ a hierarchical structure for categorizing products; however, few studies have leveraged this hierarchical information for classification. Furthermore, studies that consider hierarchical information fail to account for similarities and differences across various hierarchical categories. 
Herein, we introduce a large-scale hierarchical dataset collected from the JD e-commerce platform ({\it www.JD.com}), comprising 1,011,450 products with titles and a three-level category structure. By making this dataset openly accessible, we provide a valuable resource for researchers and practitioners to advance research and applications associated with product categorization. Moreover, we propose a novel hierarchical text classification approach based on the widely used Bidirectional Encoder Representations from Transformers (BERT), called Hierarchical Fine-tuning BERT (HFT-BERT). HFT-BERT leverages the remarkable text feature extraction capabilities of BERT, achieving prediction performance comparable to those of existing methods on short texts. Notably, our HFT-BERT model demonstrates exceptional performance in categorizing longer short texts, such as books.
\end{abstract}




\end{frontmatter}


\section{Introduction}
\label{sec:intro}

Product categorization plays a crucial role in ensuring effective functioning of e-commerce platforms \cite{zhang2020commodity,dhote2020hybrid}. It involves categorizing products based on their nature, composition, and intended use, allowing for efficient search and filtering mechanisms. This categorization process is the backbone of product organization \cite{fang2023commodity}. Through an effective classification mechanism, consumers can easily find what they need, while sellers can effectively showcase their products. However, existing e-commerce platforms heavily rely on manual product annotation \cite{hasson2021category}. According to the rules of various e-commerce platforms for adding products, sellers must manually select or fill in product categories \cite{Helpcenter}. However, the manual approach has low efficiency and consistency because various sellers might categorize the same product differently \cite{zhang2020commodity}.

\subsection{Problem Statement}

In e-commerce platforms, sellers are required to choose the most appropriate product categories from the numerous categories available \cite{chen2019fine}. Therefore, scholars have adopted text classification methods to classify products based on their titles. 
Specifically, Lee et al. \cite{lee2018engineering} adapt the doc2vec algorithm that implements the document embedding technique to develop an automatic product classifier. Chen et al. \cite{chen2019fine} propose a neural product categorization model to identify fine-grained categories from the product content. A character-level convolutional embedding layer is used to learn compositional word representations, and a spiral residual layer is used to extract word context annotations that capture complex long-range dependencies and structural information. \cite{li2019customs} propose a text-image adaptive convolutional neural network to effectively utilize website information and facilitate the customs classification process. Later, Zhang et al. \cite{zhang2020commodity} propose a commodity text-classification-based e-commerce category and attribute mining method, which takes into account the attributes of the corresponding categories between different platforms. Recently, Fang et al. \cite{fang2023commodity} use BERT to obtain text feature representations and a Variational Auto-Encoder (VAE) to address title discrimination and unbalanced sample sizes. This integrated approach is referred to as BERT-VAE.

Notably, products on e-commerce platforms often have a hierarchical category structure \cite{kozareva2015everyone}. 
To the best of our knowledge, most existing literature utilizes the technique of flat classification, ignoring the hierarchical structure \cite{zhang2016dependency,Liu2017Deep}. Recently, methods have been proposed for hierarchical classification, enhancing the predictive ability of text classification models \cite{GARGIULO2019125,gong2020hierarchical,ma2022hybrid,song-etal-2023-peer}. However, many existing studies focus on utilizing direct parent--child relationships in the hierarchy, ignoring differences and similarities in the same hierarchy level \cite{song-etal-2023-peer}. Further, most applications of the existing hierarchical text classification (HTC) methods deal with English texts, such as the RCV1 dataset \cite{lewis2004rcv1} and the Amazon670K dataset \cite{leskovec2014snap}. To the best of our knowledge, few hierarchical classification methods have been developed for Chinese texts. Therefore, developing a method that can make the full use of hierarchical information and is suitable for various hierarchical structures as well as Chinese texts is crucial. 

\subsection{Research Objective}

To make full use of hierarchical information, the Hierarchical Fine-Tuning based Convolutional Neural Network (HFT-CNN) proposed by Shimura et al. \cite{shimura2018hft} effectively utilizes the hierarchical information of categories and achieves better prediction results. Data in the upper levels are used for categorization in the lower levels by applying a CNN with a fine-tuning technique. Inspired by them \cite{shimura2018hft}, we use the BERT model instead of a CNN for the hierarchical classification of product texts. Because of the better performance of BERT in text feature extraction and Chinese text learning, we aim to achieve more accurate classification results through this replacement. Using the pretrained model of BERT, we can learn contextual information in product texts and apply it for hierarchical classification. Compared with CNN, BERT can better capture semantics and contextual information in texts, improving classification accuracy. We fine-tune the BERT model to adapt it to the hierarchical classification of product texts using its pretrained weights as the starting point. With this improvement, we obtain better prediction results and enhance the performance of hierarchical classification of product texts.

\subsection{Contribution}

Based on the above discussions, we present the contributions of our paper from three perspectives. First, we collect a large-scale Chinese dataset of products from the JD e-commerce platform ({\it www.JD.com}), which comprises 1,011,450 products. This dataset includes product titles and category names at three hierarchical levels in Chinese. There is no space between words and no punctuation mark after word endings in the Chinese text. Moreover, the number of high-frequency words in Chinese is substantially larger than that in English. Therefore, the analysis of the Chinese text is more difficult \cite{8880595}. Existing methods mostly applied to English text, while our work is based on Chinese text classification. In addition, by making this dataset open-source ({\it https://gitee.com/KunLiu\_kk/jd-dataset}), we provide a valuable resource for researchers and practitioners to advance research and applications associated with product categorization. Second, we propose a novel approach for HTC that incorporates the extensively used BERT model and utilizes hierarchical information in the dataset, referred to as Hierarchical Fine-Tuning BERT (HFT-BERT). This approach leverages the superior text feature extraction capabilities of BERT, achieving prediction performance comparable to those of existing methods on our dataset. Third, the HFT-BERT model shows short-text classification performance comparable to those of other existing models and achieves excellent prediction performance on longer short texts, e.g., book introductions, compared to those of other existing models. Thus, our model can provide support for more efficient operations on Chinese e-commerce platforms.

\section{Related Work}
\label{sec:review}

HTC plays an important role in text classification, where documents are organized in a hierarchical structure. In the early years, the most commonly employed technique to solve the HTC problem is the local classifier approach, which mainly constructs a hierarchy of traditional classifiers \cite{vateekul2014hierarchical}. According to Silla et al. \cite{silla2011survey}, the local classifier approach can be further categorized into three standard methods: training a local classifier per node (LCN), training a local classifier per parent node (LCPN), and training a local classifier per level (LCL). In contrast to the local classifier approach, the global classifier approach first treats HTC problem as a flat multiclassification problem and uses a single classifier \cite{zhou2020hierarchy}. Hierarchical information is introduced into the global classifier approach using special parameter initialization, regularization term, or model structure. Methods developed for multi-label text classification problems can be subsequently used \cite{8879472,9169885,10440286}.

\subsection{Traditional Classifier}
In the LCN approach, a classifier is trained for each category in the hierarchy to determine whether an item should be classified into this category. The classifiers for each category can be the same or different. For instance, D'Alessio et al. \cite{d2000effect} utilized 
a one-of-$M$ classifier at the top level, along with binary classifiers for each category in sub trees. To make better use of the parent--child relationship in the hierarchy, Sun et al. \cite{sun2001hierarchical} further built a support vector machine (SVM) classifier for each category and a binary classifier for each parent category. In contrast to considering the tree structure, Nguyen et al. \cite{nguyen2005text} proposed a technique based on directed acyclic graphs, where a child category may have multiple parent categories. Here, the SVM is used as a binary classifier for each category. Thus, novel structures, new classifiers, and algorithms continuously emerge in relation to the LCN approach \cite{esuli2006treeboost,duwairi2011hierarchical,he2014hierarchical}.

In the LCPN approach, each parent category is associated with a multilabel classifier or a series of binary classifiers to determine which child category or categories an item should be classified into \cite{silla2011survey}. Koller et al. \cite{koller1997hierarchically} pioneered an HTC solution using the LCPN approach, where Naive Bayes was selected as the baseline classifier for each parent category. Wu et al. \cite{wu2005learning} utilized the C4.5 decision tree as the base multilabel classifier for each parent category. Further, Moskovitch et al. \cite{moskovitch2006multiple} employed centroid classifiers to determine whether an item should be assigned to corresponding categories based on a threshold. Typically, the same classification algorithm is used throughout the hierarchy in the LCPN approach. However, Secker et al. \cite{secker2007experimental} used different classification algorithms at different parent nodes of the class hierarchy to improve prediction accuracy. Secker et al. \cite{secker2010hierarchical} further extended the approach proposed in 2007 \cite{secker2007experimental} by selecting different classifiers as well as attributes at each step when choosing classifiers.

In the LCL approach, a multilabel classifier is trained for each level in the hierarchy to determine which category or categories at that level an item should be classified into \cite{silla2011survey}. Clare et al. \cite{clare2003predicting} were the first to employ the LCL method and used the C4.5 decision tree as their multilabel classifier. Chen et al. \cite{chen2005hierarchical} used the back propagation (BP) learning model to construct appropriate hierarchical classification units for each level, which constitutes an LCL approach. In this approach, the results from top level trigger the BP classifiers in the next level, and so on. 

Unlike the local classifier, the global classifier uses a single classifier, treating the HTC problem as a flat multiclassification problem with special hierarchical information. The global classifier approach is also referred to as the Big-Bang approach \cite{silla2011survey}. Different traditional classifiers are used as the multilabel global classifiers \cite{gaussier2002hierarchical,dekel2004large,silla2009global}. In addition to using different global classifiers, several studies focus on optimizing models and improving the effectiveness of hierarchical classification. Kiritchenko et al. \cite{kiritchenko2006learning} were the first to incorporate the boosting algorithm into HTC methods, presenting a novel global classifier approach based on AdaBoost, which achieved consistent classification results. Khan et al. \cite{khan2014novel} proposed a novel ant-colony-optimization-based single-path hierarchical classification algorithm, which was further extended in their later research \cite{khan2017ant}. 

\subsection{Deep Learning Classifier}
Deep learning algorithms have attracted considerable attention since their emergence and have been widely applied in various fields particularly in text analysis. Recently, there has been a surge in research exploring the applications of deep learning algorithms to sovle HTC problems. Lecun et al. \cite{lecun2015deep} demonstrated the effectiveness of deep learning models in automatically learning the hierarchical representations of image data, leading to the widespread adoption of CNNs in the field of HTC \cite{peng2018large,GARGIULO2019125}. These CNN-based global approaches utilize graph convolution operations to process texts represented as a graph-of-words. In addition to the CNN model, other deep learning models such as encoder and decoder models have also been introduced into HTC methods to extract textual and label semantics \cite{gong2020hierarchical,chen2021hierarchy,ma2023led,ZHANG2024112153}. However, the training models for global classifier approaches are logically complicated and often suffer from underfitting owing to lost priori information about categories and their structural relationships \cite{kowsari2017hdltex}. In contrast, local classifier approaches can make better use of hierarchical information. Therefore, many studies combine deep learning with local classifier approaches.

Regarding local classifier approaches, Kowsari et al. \cite{kowsari2017hdltex} were the first to propose a local HTC approach called HDLTex based on deep learning models, utilizing Deep Neural Networks (DNNs), CNNs, and Recurrent Neural Networks (RNNs) as multilabel classifiers. The HDLTex model builds a multilabel classifier for elements belonging to the same parent category. The authors constructed a hierarchical two-level taxonomy dataset of 46,985 published papers obtained from the Web of Science {\it www.webofscience.com}. The results showed that RNNs exhibited the best classification accuracy, followed by the CNN model. Based on transfer learning methods, Banerjee et al. \cite{banerjee2019hierarchical} proposed HTrans, where binary classifiers at lower levels in the hierarchy are initialized using the parameters of the parent classifier and fine-tuned during the child-category classification task. Transfer learning methods seem to offer better performance in solving HTC problems. Compared with the attention-based gated recurrent unit (GRU) model, the HTrans method offers significant improvements of 1\% concerning micro-F1 and 3\% concerning macro-F1 on RCV1. However, the above-mentioned methods involve a huge number of parameters and computational overhead because they build or fine-tune deep learning models for almost every node \cite{sinha2018hierarchical}. Wang et al. \cite{WANG2022108576} propose a novel hierarchical classification method based on graph learning model is proposed to learn the graph embedding that well captures the node, relation, and graph structure information for hierarchical classification.

Among various approaches that utilize deep learning, the LCL approach has gained significant popularity. In the LCL approach, deep networks are constructed or fine-tuned locally for each level within the hierarchy. This strategy effectively addresses the challenges of excessive parameters and computational overhead. Shimura et al. \cite{shimura2018hft} proposed the HFT-CNN, a CNN-based LCL method with a fine-tuning technique. Results based on the RCV1 dataset \cite{lewis2004rcv1} and the Amazon670K dataset \cite{leskovec2014snap} prove the competitiveness of fine-tuning in solving HTC problems. Meanwhile, Sinha et al. \cite{sinha2018hierarchical} proposed an attention-based Long Short-Term Memory (LSTM) encoder model using the WOS dataset \cite{kowsari2017hdltex}. They showed that the use of hierarchical taxonomy can provide a more robust classifier than flat classifiers. Kowsari wt al. \cite{kowsari2017hdltex} reported that the RNN model has the best effect; hence, RNNs are popular models used in the LCL approaches. Based on the RNN model, Huang et al. \cite{huang2019hierarchical} proposed Hierarchical Attention-based RNN (HARNN), an LCL approach that develops a hierarchical attention-based recurrent layer to capture associations between texts and the hierarchical structure. Similarly, Ma et al. \cite{ma2022hybrid} proposed a hybrid embedding-based text representation for hierarchical multi-label text classification (HE-HMTC), wherein a Bidirectional GRU (Bi-GRU) model is trained at each level of the category hierarchy or taxonomy in a top--down manner. The HE-HMTC model performs well on the WOS \cite{kowsari2017hdltex} and other datasets. However, the relationship between levels remains underutilized in HARNN and HE-HMTC. To share information across hierarchy levels more flexibly and effectively, Zhang et al. \cite{zhang2022hcn} designed a novel label-based attention module, which can hierarchically extract important information from the text based on labels from different hierarchy levels.

Among the LCL approaches described above, HFT-CNN \cite{shimura2018hft} is the most similar to our method. HFT-CNN utilizes data in the upper levels for categorization in the lower levels by applying a CNN model with a fine-tuning technique. In other words, HFT-CNN transfers the parameters of CNN trained from the upper to lower levels according to the hierarchical structure and fine-tunes the parameters. The basic idea of fine-tuning is to adopt a pretrained model that has been trained on a large number of texts and continue to train it on a small number of texts \cite{devlin2018bert}. As a result, the fine-tuning technique is suitable for HTC because the sample size is smaller for the lower level in the hierarchical structure. Furthermore, hierarchical dependencies between labels lead to similar parameters, making the fine-tuning of parameters effective \cite{peng2018large,KONG2022107872}. Therefore, instead of training a CNN model on the top level, we select the pretrained BERT model called BERT-base-Chinese and fine-tune the parameters on each level in the hierarchical structure from the top to bottom.

\section{HFT-BERT framework}
\label{sec:model}

\begin{figure}[!ht]
\centering
\includegraphics[width= 1\linewidth]{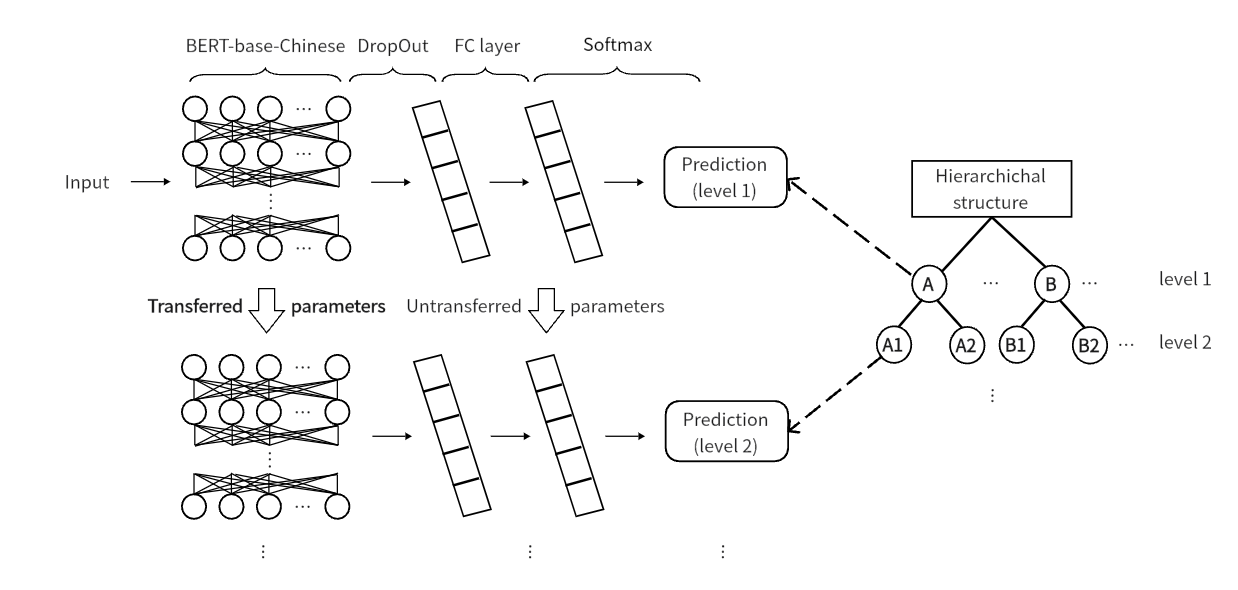}
\caption{Framework of HFT-BERT. The initial parameters of the BERT-base-Chinese model are loaded from the pretrained model and transferred and fine-tuned between different levels. Each level also has its own DropOut layer to avoid overfitting and a fully connected layer with the Softmax activation function as a classifier. }
\label{fig:HFTBERT}
\end{figure}

The framework of our model is shown in Figure \ref{fig:HFTBERT}. HFT-BERT is specifically designed for hierarchical classification, and uses the BERT-base-Chinese model as its foundation. The BERT-base-Chinese model, developed by HuggingFace team ({\it https://huggingface.co/bert}), is a pretrained model catering to simplified Chinese languages. This model is built upon a BERT-based architecture, which was initially proposed by Devlin et al. \cite{devlin2018bert}. Within our HFT-BERT framework, the initial parameters of the BERT-base-Chinese model are imported from the pretrained model, as detailed in {\it https://huggingface.co/bert}. These parameters are updated during the training process at the initial level. Then, the updated parameters at the current level are transferred to the subsequent level, where they undergo further fine-tuning during training. In addition to the BERT-base-Chinese model, each level within the HFT-BERT architecture incorporates a DropOut layer and fully connected layer with a Softmax activation function. Notably, the parameters from these two layers are also updated during training at each level within the hierarchical structure. However, they are not transferred or fine-tuned across different levels, maintaining their level-specific independence. The inclusion of the DropOut layer is a crucial step after using the BERT-base-Chinese model to avoid overfitting. Subsequently, a fully connected layer featuring a Softmax activation function is introduced as the classifier. The Softmax activation function calculates the probabilities associated with products belonging to each category within the current level. Finally, the prediction of the category of a product is determined by selecting the category with the highest probability.

To facilitate the implementation of the BERT-base-Chinese model, texts that do not reach the maximum length are padded with the special character ``\textless pad \textgreater'' during formatting. The HFT-BERT model operates sequentially, moving from the top to the bottom, indicating that it processes levels from $2$ to $3$ within the hierarchical structure of our dataset. As a result, at level $2$, categories are encoded and fed into the BERT-base-Chinese model for a 12-layer bidirectional Transformer operation, together with the word embedding vectors of the formatted texts. The parameters of the BERT-base-Chinese model are initially sourced from the pretrained BERT-base-Chinese model provided by HuggingFace team ({\it https://huggingface.co/bert}). Subsequently, the outputs generated by the BERT-base-Chinese model are directed into the DropOut layer and the fully connected layer with a Softmax activation function. We employ the cross-entropy loss function to perform gradient back propagation. Parameters are updated using the cosine-annealing learning rate schedule. During the training and evaluation phases, a batch size of 128 is employed. Upon completing the training and evaluation procedures at level $2$, the model is then assessed using the test set. The prediction accuracy for level $2$ is computed on this test set.

As HFT-BERT moves to level $3$, a noteworthy aspect is the sharing of parameters between levels $2$ and $3$. The parameters that were fine-tuned during the operation at level $2$ are loaded as the initial parameters for level $3$, and they continue to undergo fine-tuning. At this stage, categories specific to level $3$ are encoded and introduced into the BERT-base-Chinese model for a 12-layer bidirectional Transformer operation, along with the word embedding vectors of the formatted texts. Notably, the parameters within the DropOut layer and the fully connected layer are initialized separately and are not shared across levels. Following the same sequence of steps employed at level $2$, the test set is utilized, and the prediction accuracy for level $3$ is determined. 

\section{Experiments and Results}
\label{sec:experiment}

\subsection{Data Description}

We evaluate the proposed model using a real-world dataset characterized by a three-layer hierarchical structure. The dataset is obtained from JD.com, a prominent supply-chain-focused technology and service provider and a well-known comprehensive e-commerce platform in China. Our dataset comprises four categories in level $1$:{\bf fresh, household appliances, digital products, and books}. 
JD.com first engaged in electrical appliances and digital products. As a result, household appliances and digital products are the two main business categories on JD.com with several product examples. Fresh and books are the two main new product categories developed in the past decade. Due to JD Logistics, many regions can enjoy the special logistics services of the next-day or even the same-day deliveries, which is very beneficial for the sale of fresh and books. Therefore, we select these four categories in level $1$.
For all categories except books, we exclusively obtain product titles. However, because of the unique nature of books, we collect titles as well as product descriptions. For the original dataset collected by us, we perform data cleaning to remove extraneous symbols and punctuation marks from the text. Furthermore, for each category in level $1$, we randomly select 80\% data as the training set and 20\% data as the test set. An overview of our dataset is presented in Table \ref{table:dataset}. The number of categories in level $2$ is close, all very few. However, number of categories in level $3$ is large, and ``household appliances" has more categories in level $3$. In addition, the instances of ``household appliances" are the most, while those of ``books" are the least.

\begin{figure}[!ht]
\centering
\includegraphics[width=0.7\linewidth]{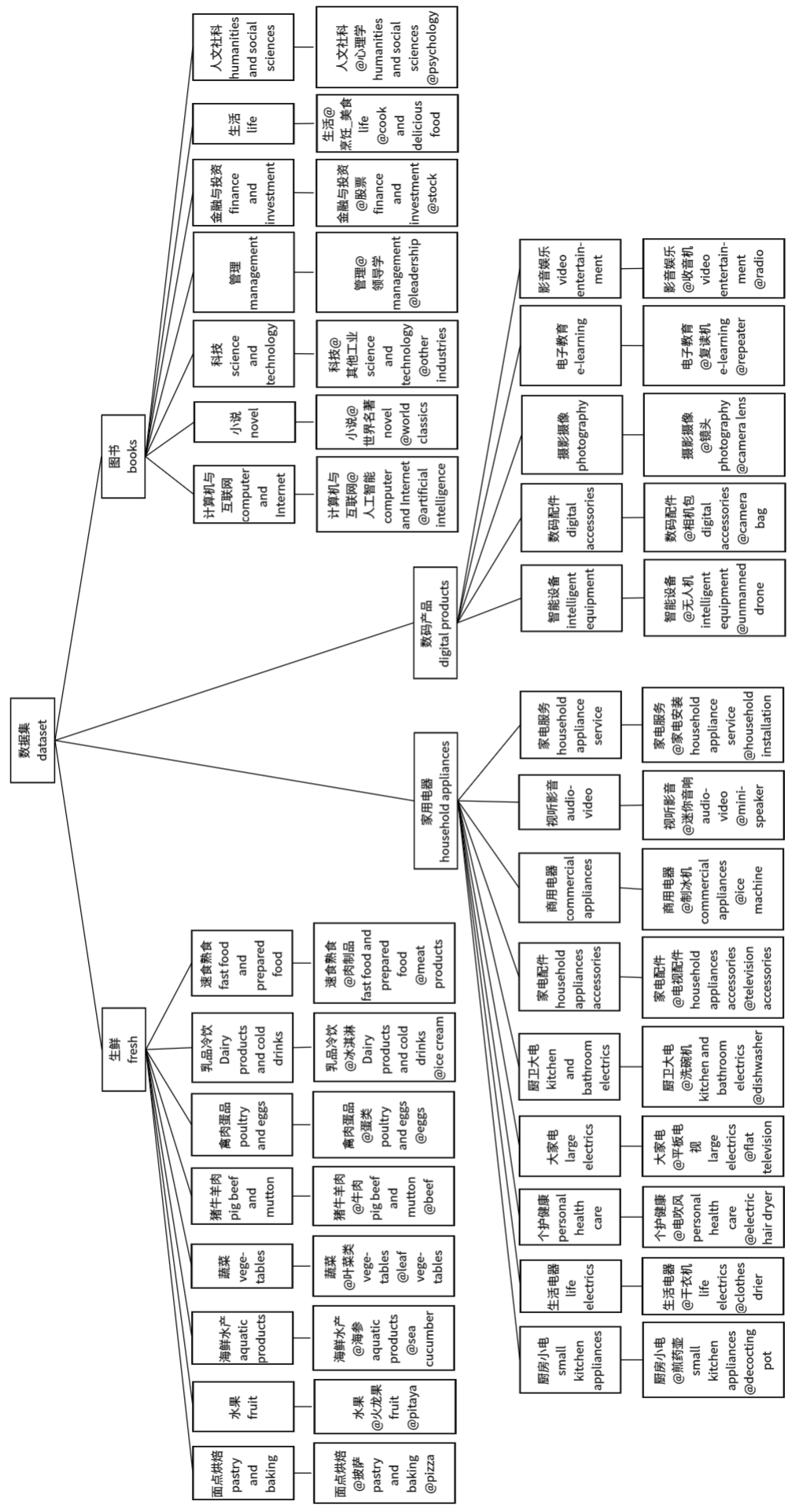}
\caption{Label tree of the dataset. Categories in level $1$ of the dataset include fresh, household appliances, digital products, and books. For each category in level $1$, all subcategories in level $2$ are shown. For each category in level $2$, only one of the subcategories in level $3$ is shown.}
\label{fig:tree}
\end{figure}

To train baseline algorithms using the code provided by Shimura et al. \cite{shimura2018hft}, we perform two key tasks: standardizing category names and constructing label trees to represent the hierarchical structure of our dataset. The categories in level $3$ are renamed in the format ``parent category in level $2$@category in level $3$". For instance, if ``novel" is a category in level $2$ and ``world classics" is a child category of ``novel", then ``world classics" is renamed as ``novel@world classics". The hierarchical structure of our dataset is visually depicted in Figure \ref{fig:tree}, which illustrates the label tree. In addition, we provide examples from our dataset in Tables \ref{table:example1}, \ref{table:example2}, \ref{table:example3} and \ref{table:example4}, showcasing one sample for each category in level $2$.

\begin{table}[!ht]
   \caption{Summary of our dataset. The first row lists the four categories in level $1$ of the dataset: fresh, household appliances, digital products, and books. We also report the number of categories in levels $2$ and $3$, the number of instances under each category in level $1$, and the number of instances in the training and test sets under each category in level $1$.}
   \vspace{0.2cm} \label{table:dataset}
   \centering
   \begin{tabular*}{\hsize}{@{}@{\extracolsep{\fill}}ccccc@{}}
      \toprule[1pt]
      Categories in level $1$ & fresh & \makecell{household \\appliances} & \makecell{digital \\products} & books \\
      \midrule
      \makecell[l]{\# of categories\\ in level $2$} & 8 & 9 & 5 & 7 \\
      \makecell[l]{\# of categories\\ in level $3$} & 80 & 172 & 50 & 85 \\
      \makecell[l]{Total sample size} & 132,175 & 580,226 & 279,116 & 19,933 \\
      \makecell[l]{sample size in the\\ training set (80\%)} & 105,740 & 464,180 & 223,292 & 15,946 \\
      \makecell[l]{sample size in the\\ test set (20\%)} & 26,435 & 116,046 & 55,824 & 3,987 \\
      \bottomrule[1pt]
   \end{tabular*}
\end{table}

\subsection{Baseline Algorithms}

We use three multilabel classifiers reported by Shimura et al. \cite{shimura2018hft} as the baselines, i.e., the Flat-CNN, Hier-CNN, and HFT-CNN models. 

\textbf{The Flat-CNN} model, as reported by Shimura et al. \cite{shimura2018hft}, is essentially a flat multilabel classifier. It treats the HTC problem as a flat multiclassification problem, ignoring the inherent hierarchical structure within the dataset. Our implementation incorporates a fastText layer for word embedding and utilizes a CNN architecture comprising three convolution layers, a max pooling layer, and a fully connected layer. To enhance model robustness, DropOut is applied within the fully connected layer.

\textbf{The Hier-CNN} model, as introduced by Shimura et al. \cite{shimura2018hft}, is a hierarchical multilabel classifier, that falls under the category of LCL approaches. The Hier-CNN approach involves the development and training of separate CNNs for each level within the hierarchy. Notably, the CNN architecture employed at each hierarchical level mirrors the  structure used in the Flat-CNN model, which comprises three convolution layers, a max pooling layer, and a fully connected layer.

\textbf{The HFT-CNN} model proposed by Shimura et al. \cite{shimura2018hft} is another hierarchical multilabel classifier, that belongs to the LCL approach. However, the primary difference between the Hier-CNN and HFT-CNN model lies in the incorporation of a fine-tuning mechanism. In the HFT-CNN model, an initial CNN is established, identical to the one employed in the Flat-CNN model, for the first hierarchical level. Subsequently, parameters derived from training of the upper levels are fine tuned at the lower levels. 

Table \ref{table:baseline} presents a comparative analysis of the three baseline algorithms along with our HFT-BERT model. We compare these four models from three aspects: whether they are flat, hierarchical, or fine-tuned. A model can only be either flat or hierarchical. The Flat-CNN is a flat model, while the other three models are hierarchical. The fine-tuning mechanism can be used for flat as well as hierarchy models. In our work, HFT-CNN as well as HFT-BERT contain fine-tuning mechanisms, whereas the other two models do not.

\begin{table}[!ht]
   \caption{Comparison of three baseline algorithms and based on HFT-BERT model over three aspects: flat, hierarchical, and fine-tuned models. The only flat model is the Flat-CNN, which is used as a contrast to the hierarchical models. Among the three hierarchical models, hier-CNN is used as a contrast to the fine-tuned models. HFT-CNN is used as a contrast to the BERT model.}
   \label{table:baseline}
   \centering
   \begin{tabular*}{\hsize}{@{}@{\extracolsep{\fill}}cccc@{}}
      \toprule[1pt]
      Model & Flat & Hierarchical & Fine-tuned \\
      \midrule
      Flat-CNN & \checkmark & / & /  \\
      Hier-CNN & / & \checkmark & / \\
      HFT-CNN & / & \checkmark & \checkmark  \\
      \textbf{HFT-BERT} & / & \checkmark & \checkmark \\
      \bottomrule[1pt]
   \end{tabular*}
\end{table}

\subsection{Results}

Experiments are conducted on the Ubuntu 18.04 operating system, which are built with the Pytorch framework, Chainer framework, CUDA 11.1 environment, and Python3.7 language. Hardware includes a GPU with NVIDIA Tesla P100-16GB and 64 GB of memory. 

Before training the models, we first perform word segmentation on the preprocessed Chinese text. Our concern is the classification of categories in levels $2$ and $3$ under each category in level $1$. As a result, we separate the dataset into four parts, each belonging to a category in level $1$, i.e., fresh, household appliances, digital products, and books. The first 30 words are reserved for commodities under the categories in level $1$, except books, and the first 200 words are reserved for books. For the baselines and our HFT-BERT model, fixed parameters include the embedding dimension of 300, and batch size of 128. 

We train the baselines and our model on our dataset. The accuracy results of the four models at level $2$ are shown in Table \ref{table:accuracy1} and the accuracy results at level $3$ are shown in Table \ref{table:accuracy2}. Flat-CNN achieves the best performance for categories fresh, household appliances, and digital products. This model ignores the inherent hierarchical structure within the dataset, predicting categories in levels $2$ and $3$ simultaneously. Therefore, it can learn more information than other models when predicting categories in level $2$, making it easier to achieve the best performance. The accuracy of our model is close to those of the baselines for categories fresh, household, and digital products. Regarding books, the accuracy of our model is 3.5\% higher than that of the best result among the three baselines and 5\% higher than those of others. 

\begin{table}[!htbp]
   \caption{Accuracy results of the four models at level $2$ under four categories at level $1$. The best performances of the four models are shown in bold. Our model achieves similar performance for categories fresh and digital products and slightly inferior performance for household appliances. Our model achieves the best performance for books.}
   \label{table:accuracy1}
   \centering
   \begin{tabular*}{\hsize}{@{}@{\extracolsep{\fill}}ccccc@{}}
      \toprule[1pt]
      model & fresh & \makecell{household\\appliances} & \makecell{digital\\products} & books \\
      \midrule
      Flat-CNN & \textbf{0.9841} & \textbf{0.9836} & \textbf{0.9854} & 0.9143 \\
      Hier-CNN & 0.9834 & 0.9827 & 0.9820 & 0.9095\\
      HFT-CNN & 0.9822 & 0.9828 & 0.9819 & 0.9008 \\
      \textbf{HFT-BERT} & 0.9816 & 0.9724 & 0.9773 & \textbf{0.9501} \\
      \bottomrule[1pt]
   \end{tabular*}
\end{table}

\begin{figure}[!ht]
    \centering
    \includegraphics[width=1\linewidth]{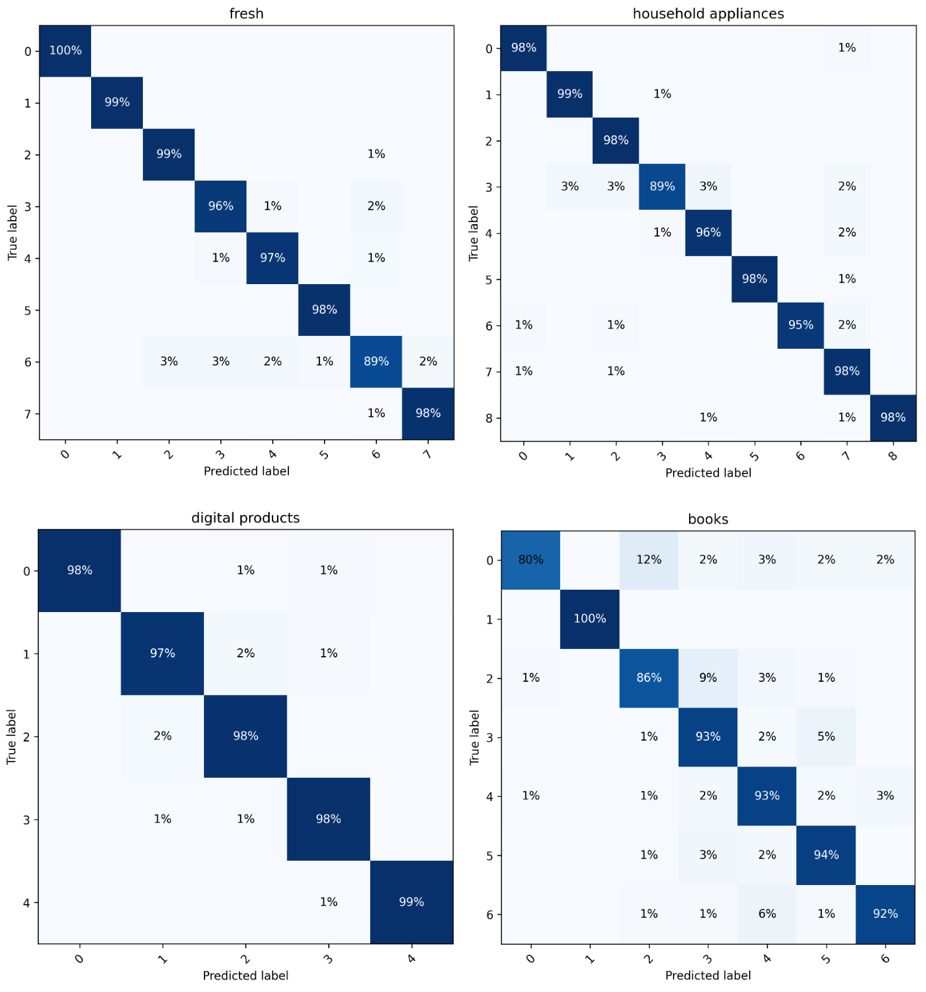}
    \caption{Confusion matrix of categories in level $2$ obtained from the HFT-BERT model. The values at the diagonal positions represent the classification accuracy of each category in level $2$. The values at the off-diagonal positions represent the misclassification rate of each category in level $2$ classified into other categories.}
    \label{fig:confusion matrixs}
\end{figure}

To present the details of the classification results, we plot the confusion matrix of the categories in level $2$ obtained by the HFT-BERT model (Figure \ref{fig:confusion matrixs}). The provided confusion matrix depicts the performance of the HFT-BERT model by presenting a square matrix with one row and one column for each category. The cells along the diagonal of the matrix represent the correctly classified instances for each category. Further, off-diagonal cells represent instances that are incorrectly classified, with each cell indicating the proportion of instances from one category that are mistakenly assigned to a different category. Through visual representation, we provide insights into the performance of HFT-BERT on longer short texts and potential areas of improvement. 

The first picture on the upper left in Figure \ref{fig:confusion matrixs} displays the classification results for fresh. The numbers 0--7 correspond to ``dairy products and cold drinks", ``fruit", ``aquatic products", ``pig beef and mutton", ``poultry and eggs", ``vegetables", ``fast food and prepared food", and ``pastry and baking", respectively. The accuracy for ``fast food and prepared food" is the lowest, which is less than 90\%. The accuracies for all other categories reach 96\% and above. Therefore, improving the classification accuracy of ``fast food and prepared food" is the key to improve the overall classification performance of the model. The low accuracy could be due to two factors: on one hand, the number of instances in ``fast food and prepared food" is the least among all categories, leading to an unsatisfactory model learning effect. On the other hand, ``fast food and prepared food" is a very vague concept: for example, frozen steaks can be classified as either ``pig beef and mutton" or ``fast food". The actual category depends on the category filled in by the merchants when encountering vague concepts, which is not standardized. 

The second picture on the upper right displays the classification result for household appliances. The numbers 0--8 correspond to ``personal health care", ``kitchen and bathroom electrics", ``small kitchen appliances", ``commercial appliances", ``large electrics", ``household appliance service", ``household appliance accessories", ``life electrics", and ``audio-video", respectively. The accuracy for ``commercial appliances" is the lowest, which is less than 90\%. The accuracies for all other categories reach 95\% and above. According to the misclassified cases of ``commercial appliances", we can infer that this is caused by an unclear concept definition. ``Commercial appliances" may be incorrectly classified as ``kitchen and bathroom electrics", ``small kitchen appliances", or ``large electrics". For example, range hoods can be classified as ``small kitchen appliances", ``kitchen and bathroom electrics", ``commercial appliances", and ``large electrics". If the commodity name does not distinguish between home use and business use, it is logically correct to classify it into either category. It also depends on the merchant who subjectively determines the true classification, leading to an incorrect classification result.

The third picture on the lower left displays the classification results for digital products. 
Digital products are classified well, with accuracies above 97\%.The numbers 0--4 correspond to ``video entertainment", ``photography", ``digital accessories", ``intelligent equipment", and ``e-learning", respectively. The last picture on the lower right displays the classification results for books. The numbers 0--6 correspond to ``humanities and social societies", ``novel", ``life", ``science and technology", ``management", ``computer and Internet", and ``finance and investment", respectively. Despite our model exhibiting the highest accuracy for books, there remains substantial scope for further refinement and optimization. Performance for ``novel" is the best, achieving 100\% accuracy. However, our model performs poorly when classifying instances of ``humanities and social societies" and ``life". For instance, 12\% of instances under ``humanities and social societies" are incorrectly classified into ``life". This is because humanities, social sciences and life are closely related concepts. Further, 9\% of the instances under ``life" are incorrectly classified into ``science and technology", which may be because technology and life are always mentioned at the same time.

The accuracy of our model at level $3$ is better on all four data sets. Our model shows an average increase of 4.5\% on fresh, 2\% on the household appliances, 1\% on the digital products and 43\% on books, offering advantages over the baselines, especially on books. Because of the excessive number of three-level categories, we do not show the confusion matrix of the classification. The performance of our model comparable with those of the other models under the first three categories (fresh, household appliances, and digital products) and superior performance for books show that our model can achieve performances comparable to those of CNN-based models on short texts and outperform CNN-based models on relatively longer short texts. 

\begin{table}[!htbp]
   \caption{Accuracy results of the four models at level $3$ under four categories at level $1$. The best performances of the four models are shown in bold. Our model achieves the best performance on all datasets. In particular, our model shows improved performance compared with those of other models for books.}
   \label{table:accuracy2}
   \centering
   \begin{tabular*}{\hsize}{@{}@{\extracolsep{\fill}}ccccc@{}}
      \toprule[1pt]
      model & fresh & \makecell{household\\appliances} & \makecell{digital\\products} & books \\
      \midrule
      Flat-CNN & 0.9331 & 0.9480 & 0.9660 & 0.4929 \\
      Hier-CNN & 0.9332 & 0.9520 & 0.9643 & 0.4744\\
      HFT-CNN & 0.9332 & 0.9482 & 0.9626 & 0.4809 \\
      \textbf{HFT-BERT} & \textbf{0.9828} & \textbf{0.9801} & \textbf{0.9814} & \textbf{0.9231} \\
      \bottomrule[1pt]
   \end{tabular*}
\end{table}

\section{Conclusion}

We introduce a novel approach for HTC using HFT-BERT. Based on the large-scale hierarchical dataset we collect, which comprises 1,011,450 products with titles and category names, our model achieves prediction performance comparable to those of existing methods. In particular, HFT-BERT demonstrates exceptional performance in categorizing books utilizing titles as well as product descriptions. Our model achieves short-text classification performance comparable to those of other existing models for fresh, household appliances, and digital products. With respect to longer short texts, it demonstrates an excellent classification accuracy of 0.9231 for books. Therefore, the proposed model offers a promising approach for enhancing the efficiency and consistency of classification operations on Chinese e-commerce platforms.
In the future, we will incorporate additional hierarchical levels into our model and compare its performance with those of state-of-the-art methods for HTC. In addition, we will expand our dataset further to include more categories.



\section*{Acknowledgments}

The research of Rui Pan is supported by the National Natural Science Foundation of China (No. 72471254), the Program for Innovation Research, the Disciplinary Funds and the Emerging Interdisciplinary Project of Central University of Finance and Economics. Feifei Wang is supported by National Natural Science Foundation of China (No.72371241), the MOE Project of Key Research Institute of Humanities and Social Sciences (22JJD110001).



\section*{Conflict of interest}

The authors declare no potential conflict of interests.

\appendix
\begin{table}
\caption{Examples of our dataset. This table includes examples of each category at level $2$ under fresh.}
\label{table:example1}
\centering
\small
\rotatebox{90}{
   
   \begin{tabular}{llll}
      \toprule[1pt]
      level 1 & level 2 & level 3 & title \\
      \midrule
       fresh &  \makecell[l]{pastry \\and baking} &  \makecell[l]{pastry and \\baking@pizza}  &  \makecell[l]{pizza pie semi-finished pizza ingredients 6/8/9\\ inches spaghetti pie artisanal pizza\dots} \\
       fresh &  fruit &  fruit@pitaya  &  \makecell[l]{domestic red pitaya (dragon fruit) 5 catty single\\ fruit 300-400g healthy light meals\dots} \\
       fresh &  aquatic products  &  \makecell[l]{aquatic products\\@sea cucumber}  &  \makecell[l]{Beijing Tongrentang's quick hair sea cucumber\\ 8g 3pcs bag dried aquatic products\dots} \\
       fresh &  vegetables  &  \makecell[l]{vegetables\\@leafy vegetables}  &  \makecell[l]{Mignon's Home home grown fresh oilseed cab-\\bage 300g pot pot vegetables Beijing\dots} \\
       fresh &  pig beef and mutton  &  \makecell[l]{pig beef and\\ mutton@beef}  &  \makecell[l]{Sanpin's four seasons Australian fatty beef rolls\\ Angus beef chops grain-Fed M3 snowflake\dots} \\
       fresh &  poultry and eggs  &  \makecell[l]{poultry and \\eggs@eggs} &  \makecell[l]{Nanyang agricultural specialties pavilion my\\ hometown farm grain eggs mountain forest\dots} \\
       fresh & \makecell[l]{dairy products\\and cold drinks} &  \makecell[l]{dairy products \\and cold drinks\\@ice cream}  &  \makecell[l]{Tianqiyipin's Ice Cream Ice Cream Chocolate\\ Crunch Ice Cream Mint Yogurt\dots} \\
       fresh &  \makecell[l]{fast food and\\prepared food} &  \makecell[l]{fast food and\\ prepared food\\@meat products}  &  \makecell[l]{BERETTA's Hungarian-style smoked salami\\ cold chain delivery cuts\dots} \\
    
      \bottomrule[1pt]
   \end{tabular}
   }
\end{table}

\begin{table}
\caption{Examples of our dataset. This table includes examples of each category at level $2$ under household appliances at level $1$. }
\label{table:example2}
\centering
\small
\rotatebox{90}{
   
   \begin{tabular}{llll}
      \toprule[1pt]
      level 1 & level 2 & level 3 & title \\
      \midrule
          
       \makecell[l]{household\\ appliances} &  \makecell[l]{small kitchen\\ appliances}  &  \makecell[l]{small kitchen \\appliances\\@decocting pot}  &  \makecell[l]{Cuckoo's a pot of a hundred ways to drink\\ decocting pot automatic Chinese medicine\dots} \\
       \makecell[l]{household\\ appliances} &  life electrics  &  \makecell[l]{life electrics\\@clothes driers}  &  \makecell[l]{clothes dryer home clothes dryer double layer\\large capacity warm air speed drying small\dots} \\
       \makecell[l]{household\\ appliances} &  \makecell[l]{personal \\health care}  &  \makecell[l]{personal health\\ care@hair dryer}  &  \makecell[l]{Pinshile‘s Vertical electric hair dryer house-\\hold hot and cold constant temperature quick\dots} \\
       \makecell[l]{household\\ appliances} &  large electrics  &  \makecell[l]{large electrics\\@flat television}  &  \makecell[l]{SHARP G70FL 4T-B70BHH5 70 inches 4K\dots} \\
       \makecell[l]{household\\ appliances} &  \makecell[l]{kitchen and \\bathroom electrics}  &  \makecell[l]{kitchen and \\bathroom electrics\\@dishwashers}  &  \makecell[l]{TOSHIBA's 13/14 sets of freestanding built-\\       in dishwashers for home use\dots} \\
       \makecell[l]{household\\ appliances} & \makecell[l]{household appliances\\accessories} & \makecell[l]{household appliances \\accessories@\\television accessories} &  \makecell[l]{Xinshengtong/Skyworth’s LCD TV Remote\\Control Panel 50X3 55E\dots} \\
       \makecell[l]{household\\ appliances} &  commercial electrics  &  \makecell[l]{commercial electrics\\@ice machine}  &  \makecell[l]{Frestec's ice machine for commercial use large\\ square ice bar milk tea drinks\dots} \\
       \makecell[l]{household\\ appliances} &  audio-video  &  \makecell[l]{audio-video\\@mini-speaker}  &  \makecell[l]{PANDA's 800 portable CD player tape\\ recorder cassette player radio\dots} \\
       \makecell[l]{household\\ appliances} &  \makecell[l]{household appliance\\ service} & \makecell[l]{household appliance \\service@household\\ installation}  &  \makecell[l]{gas cooker installation service} \\
      \bottomrule[1pt]
   \end{tabular}
   }
\end{table}

\begin{table}
\caption{Examples of our dataset. This table includes examples of each category at level $2$ under digital products at level $1$. }
\label{table:example3}
\centering
\small
\rotatebox{90}{
   
   \begin{tabular}{llll}
      \toprule[1pt]
      level 1 & level 2 & level 3 & title \\
      \midrule
  
       \makecell[l]{digital\\ products} &  intelligent equipment  &  \makecell[l]{intelligent equipment\\@unmanned drone} &  \makecell[l]{HARWAR MEGA-V8 quadcopter octocopter \\UAV (customized version)} \\
       \makecell[l]{digital\\ products}  &  digital accessories  &  \makecell[l]{digital accessories\\@camera bag} &  \makecell[l]{Aibao‘s SLR camera 30/50/100/160L moisture\\proof box office home electronics\dots} \\
       \makecell[l]{digital\\ products}  &  photography  &  \makecell[l]{photography\\@camera lens}  &  \makecell[l]{Shima’s 16mm F1.4 DC DN half-frame wide-\\angle lens Sony's A6\dots} \\
       \makecell[l]{digital\\ products}  &  e-learning  & \makecell[l]{e-learning\\@repeater}  &  \makecell[l]{English listening mp3 Walkman mp4 portable\\ learning machine repeater compact\dots} \\
       \makecell[l]{digital\\ products}  &  video entertainment  &  \makecell[l]{video enter-\\tainment@radio}  &  \makecell[l]{for Tecsun R-909 radio elderly radio full band\dots} \\
      \bottomrule[1pt]
   \end{tabular}
   }
\end{table}

\hspace{4cm}
\begin{table}
\centering
\caption{Examples of dataset. This table includes examples of each category at level $2$ under books at level $1$. }
\label{table:example4}
\rotatebox{90}{
\hspace{4cm}
\begin{tabular}{lllc}
      \toprule[1pt]
      level 1 & level 2 & level 3 & title \\
      \midrule
       books  &  \makecell[l]{computer and\\ Internet}  &  \makecell[l]{computer \\and Internet\\@artificial \\intelligence}  &  \makecell[l]{``Machine Learning for Beginners" is a must-have book on\\ machine learning, with no dizzying formulas but easy-to-\\understand analogies\dots} \\
       books  &  novel  &  \makecell[l]{novel@world\\ classics}  &  \makecell[l]{A selection of Maupassant's novels Two covers shipped at\\ random. Classic translation by Lee Yuk Min, featuring\\such masterpieces as ``The Goat's Ball", ``The Necklace",\\ and more!\dots} \\
       books  &  \makecell[l]{science and\\ technology}  &  \makecell[l]{science and\\ technology\\@other \\industries}  &  \makecell[l]{``Riveting and Welding Processing Quick Calculation" is\\ based on what riveters and welders should know and be\\ able to do as well as the basic techniques they should \\master, combining theory\dots} \\
       books  &  management  &  \makecell[l]{management\\@leadership}  &  \makecell[l]{There are three parts to this book, the first part explains\\ the concept of positive thought leadership, why the author\\ uses positive thoughts to train the self, and others\dots} \\
       books  &  \makecell[l]{finance and\\ investment}  &  \makecell[l]{finance and \\investment\\@stock}  &  \makecell[l]{The ``New Stockbrokers Quick Start" starts with the quasi\\ preparations and stock market terminology that new stock-\\brokers should do before entering the market, and then go-\\es on to talk about the\dots} \\
       books  &  life  &  \makecell[l]{life@cook and\\ delicious food} &  \makecell[l]{The world of wine is complex, not only because there are so\\ many different types of wines, but also because buying wine\\ is a highly subjective endeavor\dots} \\
       books  &  \makecell[l]{humanities\\ and social\\ sciences}  &  \makecell[l]{humanities and\\ social sciences\\@psychology}  &  \makecell[l]{The book focuses on three types of deep non-monotonic\\ cognitive change - the creation of\\ novelty, the adaptation of cognitive skills to changing envi-\\ronments, and the transformation of belief systems\dots} \\
      \bottomrule[1pt]
   \end{tabular}
   }
\end{table}

\newpage

\bibliographystyle{abbrv}
\bibliography{ref}

\end{document}